\pdfoutput=1  
\documentclass{article}

\usepackage[final]{neurips_2019}

\usepackage[utf8]{inputenc} 
\usepackage[T1]{fontenc}    
\usepackage{hyperref}       
\usepackage{url}            
\usepackage{booktabs}       
\usepackage{amsfonts}       
\usepackage{nicefrac}       
\usepackage{microtype}      

\usepackage{siunitx}
\usepackage{url}
\usepackage{etoolbox}
\usepackage{soul}
\usepackage{multirow}

\setcitestyle{numbers,square}

\renewcommand{\bfseries}{\fontseries{b}\selectfont}
\newrobustcmd{\B}{\bfseries}

\usepackage{amsfonts, amsmath, amsthm, amssymb}
\DeclareMathOperator*{\argmin}{arg\,min}

\usepackage{pgfplots}
\pgfplotsset{compat=1.14}
\usetikzlibrary{calc}
\usetikzlibrary{positioning}

\newcolumntype{C}[1]{>{\centering\let\newline\\\arraybackslash\hspace{0pt}}m{#1}}

\title{ML4CO: Is GCNN All You Need? \texorpdfstring{\\}{ }Graph Convolutional Neural Networks\\ Produce Strong Baselines For Combinatorial Optimization Problems, If Tuned and Trained Properly, on Appropriate Data}

\author{%
Amin Banitalebi-Dehkordi \\
Huawei Technologies Canada Co., Ltd. \\
\texttt{amin.banitalebi@huawei.com} \\
\And
\And
Yong Zhang \\
Huawei Technologies Canada Co., Ltd. \\
\texttt{yong.zhang3@huawei.com} \\
}

\begin{document}

\maketitle

\begin{abstract}
The 2021 NeurIPS Machine Learning for Combinatorial Optimization (ML4CO) competition was designed with the goal of improving state-of-the-art combinatorial optimization solvers by replacing key heuristic components with machine learning models. The competition's main scientific question was the following: is machine learning a viable option for improving traditional combinatorial optimization solvers on specific problem distributions, when historical data is available? This was motivated by the fact that in many practical scenarios, the data changes only slightly between the repetitions of a combinatorial optimization problem, and this is an area where machine learning models are particularly powerful at.

This paper summarizes the solution and lessons learned by the Huawei EI-OROAS team in the dual task of the competition. The submission of our team achieved the second place in the final ranking, with a very close distance to the first spot. In addition, our solution was ranked first consistently for several weekly leaderboard updates before the final evaluation. We provide insights gained from a large number of experiments, and argue that a simple Graph Convolutional Neural Network (GCNNs) can achieve state-of-the-art results if trained and tuned properly. 
\end{abstract}

\section{Introduction}
Combinatorial optimization has a wide range of applications in our day-to-day lives. Examples include: retail, manufacturing, data center management, energy systems, airline scheduling, auction design, political districting, kidney exchange, scientific discovery, ridesharing, cancer therapeutics, conservation planning, disaster response, and or even in college admission \cite{bengio2021machine, 2021aaaitutorial}. Due to its importance, it has been under extensive research and explorations especially in computer science and mathematics. This has resulted in hundreds of research papers and many successful commercial and open-source solvers available to use \cite{blum2011hybrid, ehrgott2000survey,heuberger2004inverse, achterberg2009scip,gamrath2020scip,bixby2007gurobi}.

While most combinatorial optimization solvers are presented as general-purpose, one-size-fits-all algorithms, the NeurIPS 2021 ML4CO competition focuses on the design of application-specific algorithms from historical data \cite{ml4co}. This general problem captures a highly practical scenario relevant to many application areas, where a practitioner repeatedly solves problem instances from a specific distribution, with redundant patterns and characteristics. For example, managing a large-scale energy distribution network requires solving very similar CO problems on a daily basis, with a fixed power grid structure while only the demand changes over time. This change of demand is hard to capture by hand-engineered expert rules, and ML-enhanced approaches offer a possible solution to detect typical patterns in the demand history. Other examples include crew scheduling problems that have to be solved daily or weekly with minor variations, or vehicle routing where the traffic conditions change over time, but the overall transportation network does not.

The ML4CO competition  \cite{ml4co} features three challenges for ML, each corresponding to a specific control task arising in the open-source solver SCIP \cite{gamrath2020scip}, and exposed through a unified OpenAI-gym API based on the Python library Ecole \cite{ecole}. For each challenge, participants were evaluated on three problem benchmarks originating from diverse application areas, each represented as a collection of mixed-integer linear program (MILP) instances.

This paper summarizes the solution and lessons learned by the EI-OROAS team in the dual task of the competition. The submission of our team achieved the second place in the final ranking, with a very close distance to the first spot. In addition, our solution was ranked first consistently for several weekly leaderboard updates before the final evaluation. We provide insights gained from a large number of experiments, and argue that a simple Graph Convolutional Neural Network (GCNNs) can achieve state-of-the-art results if trained and tuned properly, on the right data. In addition, we provide a series of remarks and guidelines which we believe will be useful for practitioners to pay attention to, in order to get to the full potential of their NN models used for combinatorial optimization.

\section{Background}
Most combinatorial optimization problems in practice can be written as mixed-integer linear programs (MILP). Branch-and-bound (B\&B) algorithm \cite{bb} is an exact method for solving MILPs. When interrupted before termination (or completion), B\&B can provide intermediate solutions along with optimality bounds (even without actually knowing the optimal values) \citep{bb,gcn}. This algorithm builds a solution tree and recursively selects variables for branching (LP relaxation). This process was traditionally done with highly tuned hard-coded heuristics, however, over the past several years several machine learning (ML) approaches are also emerging \citep{gcn, hybrid, nair, zarpellon2021parameterizing, huang2022learning}. The idea is that for repetitive type of problems such as production planning, one may be able to use an ML model for processes such as variable selection (a.k.a. branching), in order to minimize the total solving time of MILP instances. 

A mixed-integer linear program (MILP) can be expressed as follows:
\begin{multline}
    \label{eq:milp-def}
    \hspace{150pt} \argmin_{\mathbf{x}} \mathbf{c}^\top \mathbf{x} \\ \text{s.t.} \hspace{5pt} \mathbf{A} \mathbf{x} \le \mathbf{b}, \enskip \mathbf{l} \leq \mathbf{x} \leq \mathbf{u}, \enskip \mathbf{x} \in \mathbb{Z}^{p} \times \mathbb{R}^{n-p} \text{,} \hspace{90pt} 
\end{multline}
where $\mathbf{c} \in \mathbb{R}^n$ is the objective coefficient vector, $\mathbf{A} \in \mathbb{R}^{m \times n}$ is the constraint coefficient matrix, $\mathbf{b} \in \mathbb{R}^m$, $\mathbf{l}, \mathbf{u} \in \mathbb{R}^n$, and $p \leq n$ is the number of integer variables. Moreover, $m$ is the number of rows and $n$ refers to the number of columns. 

Relaxing the integrality constraint turns the problem into a linear program (LP) which can be solved (e.g. via the simplex algorithm) to obtain a lower bound to (\ref{eq:milp-def}). Solution to the LP relaxation will be also a solution to (\ref{eq:milp-def}), assuming it still respects the original integrality constraint. Otherwise, two sub-problems can be created on the two sides (floor and ceil) of an integer variable. This binary decomposition is repeatedly done via the branch-and-bound algorithms.

Fractional variable selection in branch-and-bound can be done in different ways. These methods will have a different impact on the size of the B\&B tree, and therefore in the speed and performance of a solver \cite{ACHTERBERG200542}. Example methods include: 
\begin{itemize}
    \item strong branching (SB) \cite{tsp}: results in smallest search tree but may be slow on iterations, as it will have to solve two LPs for each branching candidate.
    \item pseudo-cost: estimates the branching effect based on variable history \cite{pseudo}
    \item hybrid: to perform SB at the beginning and slowly switch to pseudo-cost or other simpler heuristics \cite{hybrid,hybrid2}
\end{itemize}

Machine learning based solutions treat the variable selection and branching processes as a markov decision process, and try to employ learning algorithms as a replacement for conventional branching rules. Next we will briefly mention some of these ML-based strategies.

\begin{figure}[t]
  \begin{minipage}[c]{0.5\textwidth}
        \centering
        \begin{tikzpicture}[scale=0.6]
        \tikzset{font=\tiny}
        \coordinate (leftstate) at (-3.5, 0);
        \coordinate (rightstate) at (3.5, 0);
        
        \draw[thick,rounded corners] ($(-3.2, -2.6)+(leftstate)$) rectangle ($(2.2, 0.5)+(leftstate)$) {};
        \node[scale=1.2]  at ($(-0.5, 0.9)+(leftstate)$) {$\mathbf{s}_t$};
        \node [circle,draw] (m) at (leftstate) {};
        \node [circle,draw] (m1) at ($(-1, -1.1)+(leftstate)$) {};
        \node [above left = 1pt and -10pt of m1] {$x_7 \leq 0$};
        \path (m) edge[-] (m1);
        \node [circle,draw] (m0) at ($(1, -1.1)+(leftstate)$) {};
        \node [right = 0.1in of m0] {};
        \node [above right = 1pt and -10pt of m0] {$x_7 \geq 1$};
        \path (m) edge[-] (m0);
        \node [circle,draw] (m11) at ($(-2, -2.2)+(leftstate)$) {};
        \node [above left = 1pt and -10pt of m11] {$x_1 \leq 2$};
        \path (m1) edge[-] (m11);
        \node [circle,draw,fill=red!20] (m10) at ($(0, -2.2)+(leftstate)$) {};
        \node [above right = -1pt and -10pt of m10] {$x_1 \geq 3$};
        \path (m1) edge[-] (m10);
        \node [scale=1.2] at ($(-0.5, -3.1)+(leftstate)$) {$\mathcal{A}(\mathbf{s}_t)=\{1, 3, 4\}$};
        
        \path ($(2.3, -2)+(leftstate)$) 
                edge[->,thick,bend right] 
                node [below right= 6pt and -18pt,midway,scale=1.2] {$\mathbf{a}_t = 4$} 
                ($(-3.5, -3)+(rightstate)$);

        \draw[thick,rounded corners] ($(-3.25, -3.7)+(rightstate)$) rectangle ($(2.55, 0.5)+(rightstate)$) {};
        \node[scale=1.2] at ($(-0.5, 0.9)+(rightstate)$) {$\mathbf{s}_{t+1}$};
        \node [circle,draw] (n) at (rightstate) {};
        \node [circle,draw] (n1) at ($(-1, -1.1)+(rightstate)$) {};
        \node [above left = 1pt and -10pt of n1] {$x_7 \leq 0$};
        \path (n) edge[-] (n1);
        \node [circle,draw,fill=red!20] (n0) at ($(1, -1.1)+(rightstate)$) {};
        \node [right = 0.1in of n0] {};
        \node [above right = 1pt and -10pt of n0] {$x_7 \geq 1$};
        \path (n) edge[-] (n0);
        \node [circle,draw] (n11) at ($(-2, -2.2)+(rightstate)$) {};
        \node [above left = 1pt and -10pt of n11] {$x_1 \leq 2$};
        \path (n1) edge[-] (n11);
        \node [circle,draw] (n10) at ($(0, -2.2)+(rightstate)$) {};
        \node [above right = -1pt and -10pt of n10] {$x_1 \geq 3$};
        \path (n1) edge[-] (n10);
        \node [circle,draw] (n100) at ($(-1, -3.3)+(rightstate)$) {};
        \node [above left = -1pt and -10pt of n100] {$x_4 \leq -2$};
        \path (n10) edge[-] (n100);
        \node [circle,draw] (n101) at ($(1, -3.3)+(rightstate)$) {};
        \node [above right = -1pt and -10pt of n101] {$x_4 \geq -1$};
        \path (n10) edge[-] (n101);
        \end{tikzpicture}%
  \end{minipage}\hfill
  \begin{minipage}[c]{0.4\textwidth}
        \caption{A demonstration of the B\&B variable selection. Left: a node is selected by the solver to be branched in the next iteration. Right: a new state formed. Visualization from \cite{gcn}.}
        \label{fig:BnB-mdp}
  \end{minipage}
\end{figure}
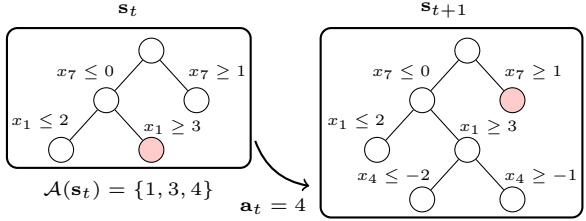

\section{Related Work}
Earlier works include methods that try to learn to mimic the strong branching rule. \cite{RePEc} defines a regression problem (predicting the SB scores) while \cite{hansknecht2018cuts,khalil2016learning} approach it as a ranking problem (predicting the order of the candidates). 
\cite{gcn} models the branching as a classification problem were labels are branching decisions from the SB rule. This GCNN-based approach for the first time reported results better than a solver (SCIP) that uses presolving, primal heuristics, and cuts.
Later, \cite{nair2020solving} built on the GCNN method by incorporating an ADMM-based expert to scale-up the full strong branching to large instances.
Other works include \cite{zarpellon2021parameterizing} where authors show that using the entire B\&B tree can further boost the imitation learning, or \cite{hybrid} where the imitation learning was made faster by switching to a small MLP after the root node of the tree.

Our solution employs the GCNN approach of \cite{gcn}. In particular, we show that the inherent incorporation of graph convolutional neural networks for this problem is a suitable choice and with proper tuning and training on suitable training data, it can still achieve state-of-the-art results.

\section{The ML4CO Competition}
\paragraph{Challenges:} 
There are three distinct challenges, each corresponding to a specific control task arising in traditional solvers: primal, dual, and configuration. This article discusses the details of the submission by the Huawei EI-OROAS team on the dual task, that is about selecting branching variables, in order to minimize the dual integral over time.

\begin{figure}
    \centering
    \includegraphics[width=0.70\linewidth]{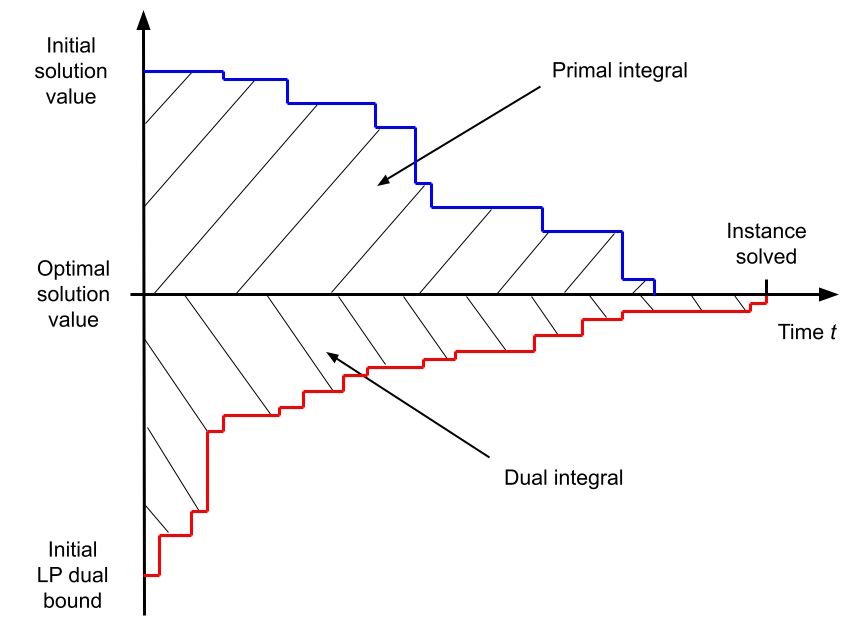}
    \caption{Primal and dual bounds evolution vs solving time. Image from \cite{ml4co}}
    \label{fig:primal_dual_integral}
\end{figure}

\paragraph{The dual task:}
The dual task deals with obtaining tight optimality guarantees (dual bounds) via branching \cite{ml4co}. Making good branching decisions is regarded as a critical component of modern branch-and-bound solvers, yet has received little theoretical understanding to this day \cite{lodi2017learning}. In this task, the environment will run a full-fledged branch-and-cut algorithm with SCIP, and the participants will only control the solver's branching decisions. The metric of interest is the dual integral, which considers the speed at which the dual bound increases over time (See Figure \ref{fig:primal_dual_integral}). Also, all primal heuristics will be deactivated, so that the focus is only on proving optimality via branching.

\paragraph{Environment:}
Environment is a traditional branch-and-bound algorithm. The solver stops after each node is processed (LP solved), receives a branching decision, performs branching, and selects the next open node to process. The solver remains in the SOLVING stage until the episode terminates (time limit of 15 minutes reached, or when the problem is solved). Action is one of the current node's branching candidates (non-fixed integer variables). Only single-variable branching is allowed.

\paragraph{Dual integral:}
This objective measures the area over the curve of the solver's dual bound (a.k.a. global lower bound), which usually corresponds to a solution of a valid relaxation of the MILP. By branching, the LP relaxations corresponding to the branch-and-bound tree leaves get tightened, and the dual bound increases over time. With a time limit $T$, the dual integral expresses as:

\begin{equation}
    Tc^\intercal x^\star - \int_{t=0}^{T} z^\star_t \,dt
\end{equation}
where $z^\star_t$ is the best dual bound at time $t$, and $Tc^\intercal x^\star$ is an instance-specific constant that depends on the optimal solution value $c^\intercal x^\star$. The dual integral is to be minimized and takes an optimal value of 0.

\paragraph{Datasets:}
There are three problem benchmarks from diverse application areas. A problem benchmark consists in a collection of mixed-integer linear program (MILP) instances in the standard MPS file format. The first two problem benchmarks are inspired by real-life applications of large-scale systems at Google, while the third benchmark is an anonymous problem inspired by a real-world, large-scale industrial application \cite{ml4co}.

\begin{itemize}
    \item Problem benchmark 1: Balanced Item Placement. This problem deals with spreading items (e.g., files or processes) across containers (e.g., disks or machines) utilizing them evenly. Items can have multiple copies, but at most, one copy can be placed in a single bin. The number of items that can be moved is constrained, modeling the real-life situation of a live system for which some placement already exists. Each problem instance is modeled as a MILP, using a multi-dimensional multi-knapsack formulation. This dataset contains 10000 training instances (pre-split into 9900 train and 100 valid instances).
    \item Problem benchmark 2: Workload Apportionment. This problem deals with apportioning workloads (e.g., data streams) across as few workers (e.g., servers) as possible. The apportionment is required to be robust to any one worker's failure. Each instance problem is modeled as a MILP, using a bin-packing with apportionment formulation. This dataset contains 10000 training instances (pre-split into 9900 train and 100 valid instances).
    \item Problem benchmark 3: Anonymous Problem. The MILP instances corresponding to this benchmark are assembled from a public dataset, whose origin is kept secret to prevent cheating. Reverse-engineering for the purpose of recovering the test set is explicitly forbidden. This dataset contains 118 training instances (pre-split into 98 train and 20 valid instances).
\end{itemize}

\section{Our Solution}
In this section, we first provide an overview of the Graph Convolutional Neural Network approach adopted from \cite{gcn}, then we provide results of our experiments, ablations, and insights around them.

\subsection{Graph Convolutional Neural Network (GCNN)}
The GCNN approach introduced in \cite{gcn} adopts an imitation learning approach to learn a fast approximation of strong branching (SB). This method models the branching policy as a graph convolutional neural network (GCNN), that in turn allows to leverage the bipartite graph representation of MILP problems. In this setting, a classification model is trained to predict the branching variable selection of the SB agent, hence the training loss function can be the softmax with cross-entropy.

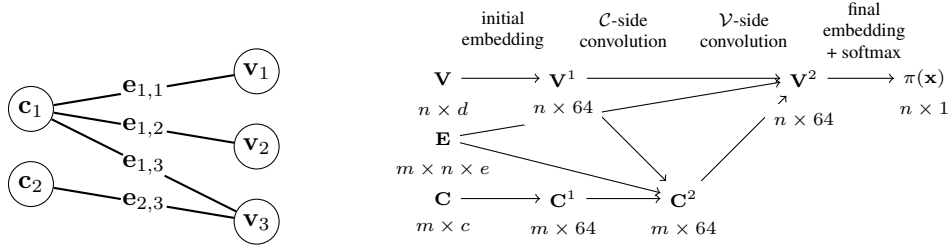
\begin{figure}[t]
\centering
\hspace{\fill}
\begin{tikzpicture}[scale=1]
    \tikzstyle{every node}=[draw,circle,minimum size=17pt,inner sep=0pt]
    \tikzstyle{annot} = [text width=6em, text centered]

    \foreach \y in {1, 2, 3}{
        \node (v\y) at (0,-\y) {$\mathbf{v}_\y$};
    }

    \foreach \y in {1, 2}{
        \path[yshift=-0.5cm, xshift=-3cm]
            node (c\y) at (0,-\y) {$\mathbf{c}_\y$};
    }

    \path (c1) edge[-, thick] node[fill=white, draw=none] {$\mathbf{e}_{1, 1}$} (v1);
    \path (c1) edge[-, thick] node[fill=white, draw=none] {$\mathbf{e}_{1, 3}$} (v3);
    \path (c1) edge[-, thick] node[fill=white, draw=none] {$\mathbf{e}_{1, 2}$} (v2);
    \path (c2) edge[-, thick] node[fill=white, draw=none] {$\mathbf{e}_{2, 3}$} (v3);

\end{tikzpicture}
\hspace{\fill}
\hspace{\fill}
\begin{tikzpicture}[scale=0.8]
\tikzset{font=\scriptsize}

\node (C0) at (0, 0) {$\mathbf{C}$};
\node (E0) at (0, 1) {$\mathbf{E}$};
\node (V0) at (0, 2) {$\mathbf{V}$};

\node (C1) at (2, 0) {$\mathbf{C}^1$};
\node (V1) at (2, 2) {$\mathbf{V}^1$};

\node (C2) at (4, 0) {$\mathbf{C}^2$};

\node (V2) at (6, 2) {$\mathbf{V}^2$};

\node (V3) at (8, 2) {$\pi(\mathbf{x})$};

\path[draw,->] (C0) -- (C1);
\path[draw,->] (V0) -- (V1);

\path[draw,->] (C1) -- (C2);
\path[draw,->] (E0) -- (C2);
\path[draw,->] (V1) -- (C2);

\path[draw,->] (C2) -- (V2);
\path[draw,->] (E0) -- (V2);
\path[draw,->] (V1) -- (V2);

\path[draw,->] (V2) -- (V3);

\node[fill=white, below=-0.01 of V0] {$n \times d$};
\node[fill=white, below=-0.01 of E0] {$m \times n \times e$};
\node[fill=white, below=-0.05 of C0] {$m \times c$};
\node[fill=white, below=-0.05 of V1] {$n \times 64$};
\node[fill=white, below=-0.05 of C1] {$m \times 64$};
\node[fill=white, below=-0.05 of C2] {$m \times 64$};
\node[fill=white, below=+0.10 of V2] {$n \times 64$};
\node[fill=white, below=-0.05 of V3] {$n \times 1$};

\node[align=center] at (1, 2.8) {initial\\embedding};
\node[align=center] at (3, 2.8) {$\mathcal{C}$-side\\convolution};
\node[align=center] at (5, 2.8) {$\mathcal{V}$-side\\convolution};
\node[align=center] at (7, 2.8) {final\\embedding\\+ softmax};

\end{tikzpicture}
\hspace{\fill}
\caption{Left: bipartite state representation $\mathbf{s}_t = (\mathcal{G}, \mathbf{C}, \mathbf{E}, \mathbf{V})$ with $n=3$ variables and $m=2$ constraints. Right: bipartite GCNN architecture for parametrizing the imitation policy $\pi$. Visualization from \cite{gcn}.}
\label{fig:bipartite}
\end{figure}

Figure \ref{fig:bipartite} shows the bipartite representation model adopted in the GCNN approach. In this figure, node and edge features are denoted by $(\mathcal{G}, \mathbf{C}, \mathbf{E}, \mathbf{V})$. Moreover, $\mathbf{C} \in \mathbb{R}^{m\times c}$, $\mathbf{V} \in \mathbb{R}^{n\times d}$, $\mathbf{E} \in \mathbb{R}^{m\times n\times e}$ respectively refer to constraints features (one per row), variable features (one per column), and edge features (sparse tensor). Table \ref{table:state-features} further shows the details of the features incorporated.

The GCNN approach utilizes two interleaved convolutions, one from variables to constraints and one from constraints to variables. In addition, the neural network used incorporates relu activations as well as prenorm layers.

\begin{table}[ht]
    \caption{Constraints, edges, and variables features used in $\mathbf{s}_t = (\mathcal{G}, \mathbf{C}, \mathbf{E}, \mathbf{V})$; adopted from \cite{gcn}.}
    \vspace{10pt}
    \label{table:state-features}
    \centering
    \begin{tabular}{c p{1.8cm} l}
    \multicolumn{1}{c}{Tensor} & \multicolumn{1}{c}{Feature} & \multicolumn{1}{c}{Description} \\
    \toprule
    \multirow{5}{*}{$\mathbf{C}$}
    & obj\_cos\_sim & Cosine similarity with objective. \\
    \cmidrule{2-3}
    & bias & Bias value, normalized with constraint coefficients. \\
    \cmidrule{2-3}
    & is\_tight & Tightness indicator in LP solution. \\
    \cmidrule{2-3}
    & dualsol\_val & Dual solution value, normalized. \\
    \cmidrule{2-3}
    & age & LP age, normalized with total number of LPs. \\
    \midrule
    \multirow{1}{*}{$\mathbf{E}$}
    & coef & Constraint coefficient, normalized per constraint. \\
    \midrule
    \multirow{13}{*}{$\mathbf{V}$}
    & type & Type (binary, integer, impl. integer, continuous) as a one-hot encoding. \\
    \cmidrule{2-3}
    & coef & Objective coefficient, normalized. \\
    \cmidrule{2-3}
    & has\_lb & Lower bound indicator. \\
    \cmidrule{2-3}
    & has\_ub & Upper bound indicator. \\
    \cmidrule{2-3}
    & sol\_is\_at\_lb & Solution value equals lower bound. \\
    \cmidrule{2-3}
    & sol\_is\_at\_ub & Solution value equals upper bound. \\
    \cmidrule{2-3}
    & sol\_frac & Solution value fractionality. \\
    \cmidrule{2-3}
    & basis\_status & Simplex basis status (lower, basic, upper, zero) as a one-hot encoding. \\
    \cmidrule{2-3}
    & reduced\_cost & Reduced cost, normalized. \\
    \cmidrule{2-3}
    & age & LP age, normalized. \\
    \cmidrule{2-3}
    & sol\_val & Solution value.  \\
    \cmidrule{2-3}
    & inc\_val & Value in incumbent. \\
    \cmidrule{2-3}
    & avg\_inc\_val & Average value in incumbents. \\
    \bottomrule
    \end{tabular}
\end{table}

\subsection{Results}
\subsubsection{Results of variable selection (classification accuracy)}
Table \ref{table:results-classification} shows the accuracy of the classification task over a number of different experiments on the item-placement dataset. In each one of these experiments, we generated a training dataset by recording the states of the MILP branching and forming a classification problem from the variable selection. Therefore, one can use different environment settings to generate different training samples set. These settings may include: time-limit to solve or terminate (in training), probability of strong branching (to use SB only on a portion of samples and pseudo-cost on the rest), number of training samples recorded (we can keep generating samples until a desired count is reached), etc.

We can see in Table \ref{table:results-classification} that in general models with higher top-k classification accuracy are somehow correlated with higher overall reward (better dual integral hence convergence), however, this relationship is not linear and thus better top-1 does not guarantee a better reward. Moreover, it is worth noting that the load-balancing and anonymous datasets also demonstrate a similar behavior, further reinforcing the conclusion made (results omitted for brevity).

\textbf{Remark \#1:} Top-1 accuracy is not enough. One needs to always check the final reward over the evaluation set (validation or test). It is also worth noting that in addition to the reward evaluated here (i.e. dual integral), there are other classical benchmarking metrics such as: ``1) shifted geometric mean of the
solving times in seconds, including running times of unsolved instances without extra penalization
(Time); 2) the hardware-independent final node counts of instances that are solved by all baselines
(Nodes); and 3) the number of times each branching policy results in the fastest solving time, over the
number of instances solved (Win)'' \cite{gcn}.

\begin{table}[ht]
    \caption{Classification accuracy for various experiments on the item-placement validation set.}
    \vspace{-4pt}
    \label{table:results-classification}
    \centering
    \begin{tabular}{c C{1.8cm} C{1.5cm} C{1.5cm} C{3.8cm}}
    \toprule
    Method & Time-Limit & P of SB & Top-1 (\%) & Avg. Accumulated Reward \\
    \toprule
    FSB & 15 min &  1.0 & - & 3771 \\
    Reliable Branching & 15 min & - & - & 3506 \\
    \toprule
    GCNN & 1 min & 1.0 & 93.9 & 6663 \\
    GCNN & 15 min & 1.0 & 98.6 & 6638 \\
    GCNN & 60 min & 1.0 & 98.9 & 6669 \\
    GCNN & 600 min & 1.0 & 99.3 & 6449 \\
    GCNN$^\star$ & 15 min & 0.001 & 67.2 & 7221 \\
    \bottomrule
    \end{tabular}
    \\ $^\star$ Refers to our best result, which will be explained later.
\end{table}

\subsubsection{GCNN results on the validation set}

Table \ref{table:results-gcnn-val} shows the results of our best trained models over the validation set problems of the three datasets. The evaluation time-limit is 15 minutes and the same setting as the test evaluations are used. Since there are 100 MILPs in the evaluation set, it takes around 24 hours for a complete evaluation on the validation set MILP instances (for the Anonymous dataset there are only 20 validation MILPs but as outlined in the competition we run them 5 times with different seeds).

For training, we used a batch size of 64 for the Item-Placement and Anonymous datasets and a batch of size 32 for the Load-Balancing (since it has larger problems that occupy more GPU memory). We trained on servers with 8$\times$V100 GPUs with 32GB memory, for 50 epochs. We utilized an Adam optimizer with learning rate of 0.001, decay on plateau (10 epochs) and early termination of 20 epochs of non-decreasing loss. Training samples set size is 100K for Item-Placement and Anonymous datasets, but only 20K for the Load-Balancing dataset as data collection is slower for this dataset.

\begin{table}[ht]
    \caption{GCNN results on the validation set MILPs.}
    \vspace{-4pt}
    \label{table:results-gcnn-val}
    \centering
    \begin{tabular}{C{2.28cm} C{1.58cm} C{1.65cm} C{1cm} C{1.1cm} C{1.3cm} C{1.41cm}}
    \toprule
    Dataset & Data Size (Train-Val) & Embedding Dimension & Time Limit & P of SB & Top-1 (\%) & Reward \\
    \toprule
    Item-Placement & 100K-1K & 8 & 15 min & 0.001 & 67.2 & 7221 \\
    Load-Balancing & 20K-1K & 64 & 1 min & 0.01 & 8.0 & 625363 \\
    Anonymous & 100K-1K & 64 & 15 min & 1.0 & 41.6 & 32517856 \\
    \bottomrule
    \end{tabular}
\end{table}

\subsubsection{Full test set evaluation results}
Table \ref{table:results-test-item}-\ref{table:results-test-anon} show the final test results of the competition, among all the submissions for the dual task. Our submission is ``EI-OROAS''. The models used for generating Table \ref{table:results-gcnn-val} were used in this submission. It can be observed that the top two submissions are very close. GCNN achieves top rank on the Load-Balancing dataset and rank 2 with a small margin on the Item-Placement dataset. It is also worth noting that after the competition deadline, we were able to achieve even better results on these two datasets. Also, in Table \ref{table:results-test-item} we only show the top 10 best results, out of the overall 23 teams \cite{ml4co}. Table \ref{table:results-test-average} shows the average ranks over the three datasets.

\textbf{Remark \#2}: GCNN results are strong. This means even though there have been papers published after the original GCNN paper \cite{gcn}, proper tuning can still keep GCNN performance on top.

\textbf{Remark \#3}: An interesting observation is that there are 16 different teams in the top 10 rewards of three datasets. This means unique team solutions were not able to achieve equally good results on different datasets, suggesting data-dependent solutions may on average be the way to go. That being said, our GCNN results are strong across all datasets, proving GCNNs are promising for these tasks.

\begin{table}[ht]
    \caption{Item-Placement top 10 test rewards from the competition website \cite{ml4co}.}
    \vspace{-4pt}
    \label{table:results-test-item}
    \centering
    \begin{tabular}{C{3cm} C{1.7cm} C{1.7cm}}
    \toprule
    Team Name & Reward & Rank \\
    \toprule
    Nuri & 6684.0 & 1 \\
    \textbf{EI-OROAS} & \textbf{6670.3} & \textbf{2} \\
    EFPP & 6487.53 & 3 \\
    lxj24 & 6443.55 & 4 \\
    ark & 6419.91 & 5 \\
    qqy & 6377.23 & 6 \\
    KAIST\_OSI & 6196.56 & 7 \\
    nf-lzg & 6077.72 & 8 \\
    Superfly & 6024.20 & 9 \\
    Monkey & 5978.65 & 10\\
    \bottomrule
    \end{tabular}
\end{table}

\begin{table}[ht]
    \caption{Load-Balancing top 10 test rewards from the competition website \cite{ml4co}.}
    \vspace{-4pt}
    \label{table:results-test-load}
    \centering
    \begin{tabular}{C{3cm} C{1.7cm} C{1.7cm}}
    \toprule
    Team Name & Reward & Rank \\
    \toprule
    \textbf{EI-OROAS} & \textbf{631744.31} & \textbf{1} \\
    KAIST\_OSI & 631410.58 & 2 \\
    EFPP & 631365.02 & 3 \\
    DaShun & 630898.25 & 4 \\
    blueterrier & 630826.33 & 5 \\
    Nuri & 630787.18 & 6 \\
    gentlemenML4CO & 630752.94 & 7 \\
    comeon & 630750.66 & 8 \\
    Superfly & 630746.96 & 9 \\
    Monkey & 630737.85 & 10\\
    \bottomrule
    \end{tabular}
\end{table}

\begin{table}[ht]
    \caption{Anonymous top 10 test rewards from the competition website \cite{ml4co}.}
    \vspace{-4pt}
    \label{table:results-test-anon}
    \centering
    \begin{tabular}{C{3cm} C{2cm} C{1.7cm}}
    \toprule
    Team Name & Reward & Rank \\
    \toprule
    Nuri & 27810782.42 & 1 \\
    qqy & 27221499.03 & 2 \\
    null\_ & 27184089.51 & 3 \\
    \textbf{EI-OROAS} & \textbf{27158442.74} & \textbf{4} \\
    DaShun & 27151426.15 & 5 \\
    KEP-UNIST & 27085394.46 & 6 \\
    lxj24 & 27052321.48 & 7 \\
    THUML-RL & 26824014.00 & 8 \\
    KAIST\_OSI & 26626410.86 & 9 \\
    Superfly & 26373350.99 & 10\\
    \bottomrule
    \end{tabular}
\end{table}

\begin{table}[ht]
    \caption{Average ranks over the three datasets for top 10 teams in the ML4CO competition \cite{ml4co}.}
    \vspace{-4pt}
    \label{table:results-test-average}
    \centering
    \begin{tabular}{C{3cm} C{3cm} C{2cm}}
    \toprule
    Team Name & Rank Multiplication & Average Rank \\
    \toprule
    \textbf{EI-OROAS} & \textbf{8} & \textbf{2.33} \\
    Nuri & 6 & 2.67 \\
    KAIST\_OSI & 126 & 6 \\
    EFPP & 117 & 6.33 \\
    qqy & 132 & 6.33\\
    DaShun & 280 & 7.67 \\
    Superfly & 810 & 9.3 \\
    lxj24 & 532 & 10\\
    comeon & 1056 & 10.3 \\
    Monkey & 1100 & 10.3 \\
    \bottomrule
    \end{tabular}
\end{table}

\subsubsection{Partial test set evaluation results prior to the final evaluation}
Throughout the ML4CO competition, leaderboard was updated on a weekly basis. GCNN results were consistently strong throughout the competition timeline. Table \ref{table:results-test-average-partial} shows the last 4 weeks of leaderboard updates for top 5 teams. Note that the evaluations were only on 25\% of the final test set.

\begin{table}[ht]
    \caption{Average ranks of the last 4 leaderboard updates over the three datasets.}
    \vspace{-4pt}
    \label{table:results-test-average-partial}
    \centering
    \begin{tabular}{C{3cm} C{1.cm} C{1.cm} C{1.cm} C{1.cm} C{3.6cm}}
    \toprule
    Team Name & Oct.04 & Oct.08 & Oct.15 & Oct.23 & Overall Average Rank \\
    \toprule
    \textbf{EI-OROAS} & \textbf{1} & \textbf{1.33} & \textbf{2.67} & \textbf{1.67} & \textbf{1.67} \\
    Nuri & 2.33 & 4.33 & 3.67 & 3.67 & 3.5 \\
    KAIST\_OSI & 8.67 & 3.67 & 6.67 & 4.67 & 5.92 \\
    EFPP & - & - & - & 5 & 5\\
    qqy & 5.67 & 4.67 & 6.67 & 8 & 6.25 \\
    \bottomrule
    \end{tabular}
\end{table}

\subsubsection{Inference speed and latency}
Capacity of a neural network is generally related to its number of parameters. Bigger models with a higher number of parameters usually lead to higher degrees of performance. For our problem of variable selection however, choosing very large architectures will be counter-productive, as these larger models come at a cost of slower inference. This means each classification within each solver iteration is slower. Sometimes this may still lead to good reward values if being accurate is more important; sometimes on the other hand speed may become a more important factor and we may be able to tolerate a certain degree of classification error. This will also be data-dependent. Smaller size problems may get solved faster, so it won't make sense to use a large NN in this setting. For harder/larger MILP instances however, it may be beneficial to choose a bigger model to push the accuracy up. Table \ref{table:models} shows the size and inference latency (averaged over 10000 runs) of the models used in our final submission. As we can see from this table, Item-Placement dataset has clearly benefited from a smaller swift model.

\textbf{Remark \#4}: We need to carefully find a balance between NN accuracy and speed. Smaller models tend to be less accurate on classification (variable selection), but because they are faster, overall they lead to better solutions (closing the dual gap faster, as long as they can learn), hence higher rewards.

\begin{table}[ht]
    \caption{DNN model size and inference latency (per node in milli-seconds on validation MILPs).}
    \vspace{-4pt}
    \label{table:models}
    \centering
    \begin{tabular}{C{2.28cm} C{1.65cm} C{1.05cm} C{1.6cm} C{1.4cm} C{1.3cm} C{1.41cm}}
    \toprule
    Dataset & Embedding Dimension & Model Size & \# parameters & Inference Latency & Top-1 (\%) & Reward \\
    \toprule
    Item-Placement & 8 & 20 {\small KB} & 1264 & 0.12 & 67.2 & 7221 \\
    Load-Balancing & 64 & 270 {\small KB} & 63872 & 9.25 & 8.0 & 625363 \\
    Anonymous & 64 & 270 {\small KB} & 63872 & 0.74 & 41.6 & 32517856 \\
    \bottomrule
    \end{tabular}
\end{table}

\subsubsection{Training and evaluation time}
Table \ref{table:training-evaluation-time} shows the time that takes to perform training for 50 epochs and evaluation on the 100 validation set MILPs on an 8-node V100 server. We can see that training is relatively fast, but evaluation takes much longer due to the time-limit of 15 minutes that is used in the competition.

\begin{table}[ht]
    \caption{Training and evaluation time (on validation set MILPs with 15 minutes solver time-limit)}
    \vspace{-4pt}
    \label{table:training-evaluation-time}
    \centering
    \begin{tabular}{C{2.28cm} C{1.65cm} C{1.65cm} C{1.6cm} C{1.8cm} C{1.6cm}}
    \toprule
    Dataset & \# Training MILPs & \# Training Samples & Training Time (hr) & \# Evaluation MILPs & Evaluation Time (hr) \\
    \toprule
    Item-Placement & 9.9K & 100K & $\approx$ 0.6 & 100 & $\approx$ 25 \\
    Load-Balancing & 9.9K & 20K & $\approx$ 1.2 & 100 & $\approx$ 25 \\
    Anonymous & 100 & 100K & $\approx$ 0.84 & 20 & $\approx$ 25 \\
    \bottomrule
    \end{tabular}
\end{table}

\subsubsection{Significant variation in performance when using a same architecture but collecting different training samples}
\begin{figure}[!b]
    \centering
    \includegraphics[width=0.70\linewidth]{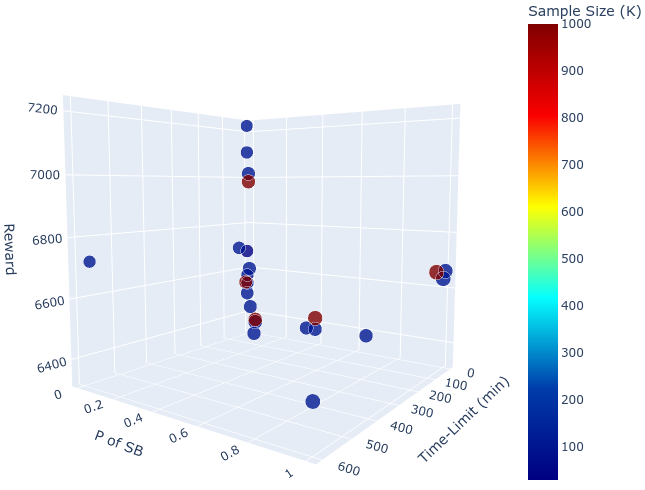}
    \caption{Validation reward in terms of solving time-limit, probability of strong branching, and training sample size used in the data collection process for different experiments (item-placement).}
    \label{fig:scatter-item}
\end{figure}

\begin{figure}[t]
    \centering
    \includegraphics[width=0.70\linewidth]{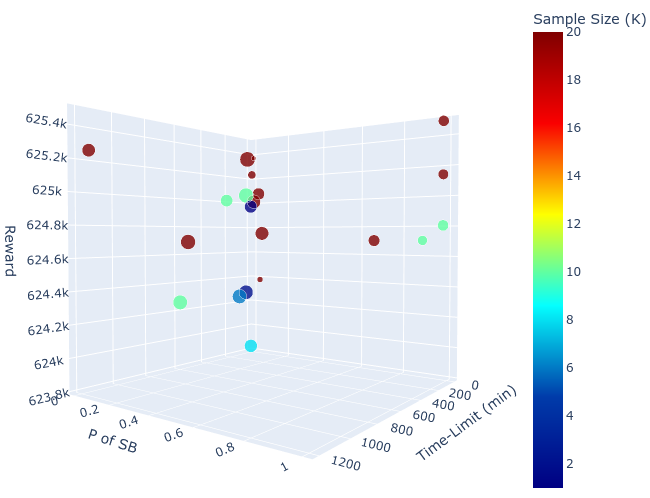}
    \caption{Validation reward in terms of solving time-limit, probability of strong branching, and training sample size used in the data collection process for different experiments (load-balancing).}
    \label{fig:scatter-load}
\end{figure}

\begin{figure}[t]
    \centering
    \includegraphics[width=0.70\linewidth]{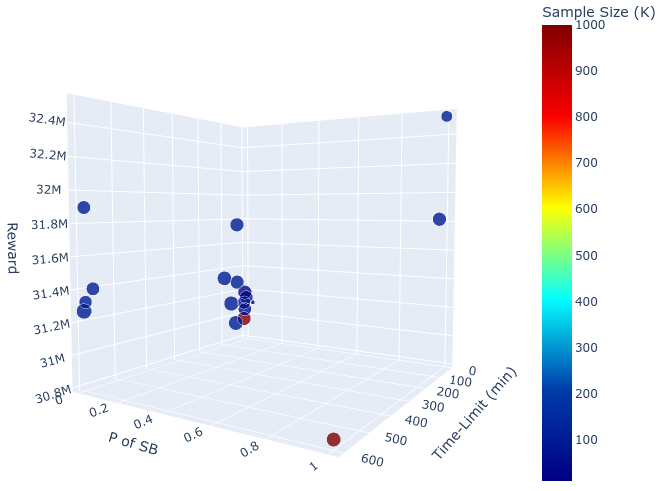}
    \caption{Validation reward in terms of solving time-limit, probability of strong branching, and training sample size used in the data collection process for different experiments (anonymous).}
    \label{fig:scatter-anon}
\end{figure}

Figure \ref{fig:scatter-item}, \ref{fig:scatter-load}, and \ref{fig:scatter-anon} show scatter plots of validation MILPs reward versus the solving time-limit, data size, and probability of strong branching, over a large number of experiments.

\textbf{Remark \#5}:
As observed, there is a significant variation in the validation reward when different flavours of training samples are collected. This means one needs to investigate various kinds of training samples settings in order to obtain the best policy.

\textbf{Remark \#6}:
Since we can collect different sets of training samples from a same set of MILP instances, a fair comparison of different methods is only possible if the original MILP instances are provided. This is the case in the ML4CO challenge. However, for problem instances such as Set Covering, Combinatorial Auction, Capacitated Facility Location, or Maximum Independent Set type problems, that are largely used in the research papers, often times it is not the case. We argue that generating different training sample sets from say Set Covering problems can lead to different final reward values. Therefore, we recommend to use a standard benchmark of MILP instances that are publicly released for such kinds of performance evaluations (e.g. datasets used in the ML4CO challenge).

\subsubsection{The impact of the training sample set size}
Figure \ref{fig:scatter-item}, \ref{fig:scatter-load}, and \ref{fig:scatter-anon} show the impact of training sample set size in the overall reward for the three datasets (see the color coding). 

\textbf{Remark \#7}: We observe that larger training set size does not necessarily result in a higher reward. This is unlike the common intuition that more training data can reduce the over-fitting.

\subsubsection{Top-1, Top-3, Top-5, and Top-10}
\begin{figure}[t]
    \centering
    \includegraphics[width=0.70\linewidth]{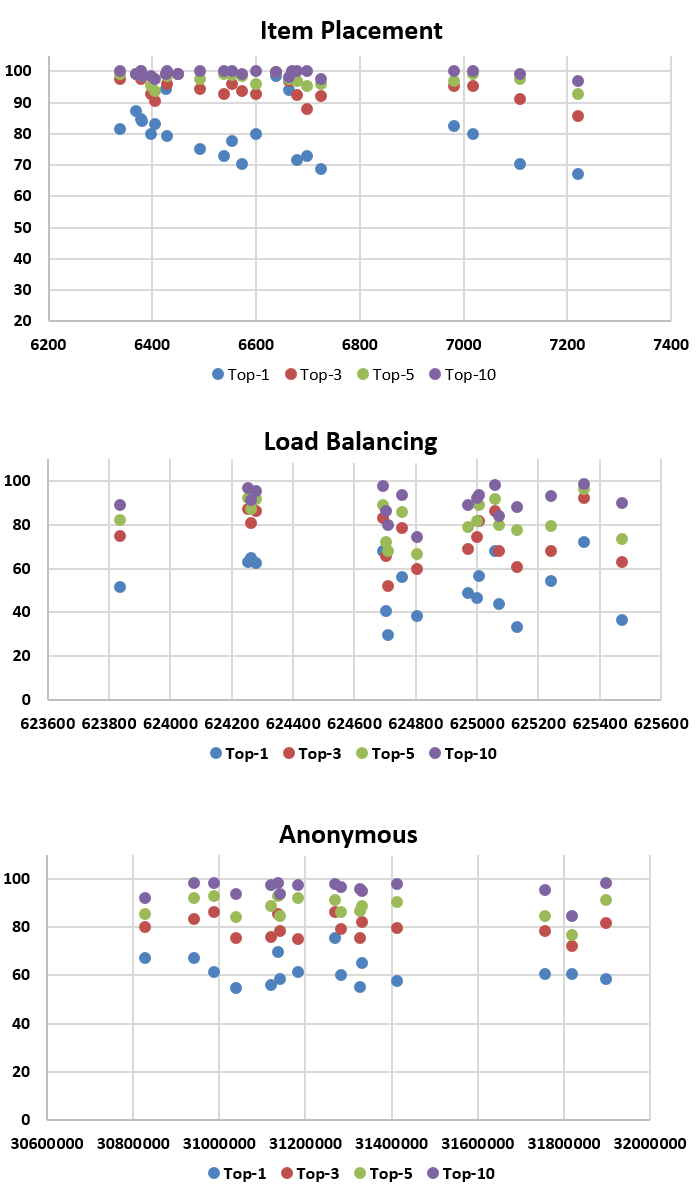}
    \caption{Validation top-1, top-3, top-5, and top-10 versus the overall reward.}
    \label{fig:topk-reward}
\end{figure}

As mentioned previously in this section, top-1 variable selection accuracy does not correlate well with the overall reward. For example, a large model with high capacity, will likely achieve a high top-1 classification accuracy, but if it gets too slow, it may end up hurting the final reward as it can do fewer number of iterations within a given time. Figure \ref{fig:topk-reward} demonstrates the reward-top-k scatter plot for the validation sets over a large number of experiments. We observe that:

\textbf{Remark \#8}: Top-5 and top-10 variable selection is in general very accurate, most of the times above 90\%. This suggests that variable selection in the branch-and-bound tree is in general fairly accurate with GCNNs, thus overall they show a strong performance compared to conventional heuristic rules.

\subsubsection{Data-Centric AI for CO}
A trend has recently been emerging in the AI community to design data-centric AI systems and tools, as opposed to only model-centric approaches \cite{dcai2021}. The idea is that the community has focused highly on building new algorithms and models, however, in practice data drift results in considerable mismatch between the training set and test/deployment environment. Based on the results of our experiments, we also conclude that data engineering plays an important role in the combinatorial optimization learning approaches, and the community needs to pay equal attention to both model/algorithm improvements and data engineering efforts.

\subsubsection{Other thoughts}
There are other ideas that we didn't try due to lack of time and resources, including: a) applying data augmentation, and b) employing multiple models. In terms of augmentation, it is important to note that one needs to take into account the nature of the data that is on linear equations when designing the augmentation operations. As for using multiple models, this idea comes from the fact that when we applied a clustering algorithm on the anonymous dataset, it revealed 3 cluster of problems, suggesting that training 3 separate GCNN models, one for each cluster of problems, may lead to performance improvements. We leave these ideas for the interested readers to explore.

\section{Conclusion}
This paper summarizes the solution and lessons learned by the Huawei EI-OROAS team in the dual task of the NeurIPS 2021 ML4CO competition. The submission of our team achieved the second place in the final ranking, with a very close distance to the first spot. In addition, our solution was ranked first consistently for several weekly leaderboard updates before the final evaluation. We provide insights gained from a large number of experiments, and argue that a simple Graph Convolutional Neural Network (GCNNs) can achieve state-of-the-art results if trained and tuned properly.

{\small
\bibliography{arxiv_main}
}

\end{document}